\documentclass[conference]{IEEEtran}
\IEEEoverridecommandlockouts
\usepackage{booktabs}

\usepackage{cite}
\usepackage{comment}
\usepackage{amsmath,amssymb,amsfonts}
\usepackage{algorithmic}
\usepackage{graphicx}
\usepackage{multirow}
\usepackage{textcomp}
\usepackage{xcolor}
\usepackage{array} 
\usepackage{amssymb}  

\usepackage{amsmath}
\usepackage{amsfonts}
\usepackage{verbatim}
\usepackage[utf8]{inputenc}

\def\BibTeX{{\rm B\kern-.05em{\sc i\kern-.025em b}\kern-.08em
    T\kern-.1667em\lower.7ex\hbox{E}\kern-.125emX}}
\begin{document}

\title{Indoor Localization using Compact, Telemetry-Agnostic, Transfer-Learning Enabled Decoder-Only Transformer}

\author{
    \IEEEauthorblockN{Nayan Sanjay Bhatia\textsuperscript{1}, Pranay Kocheta\textsuperscript{2}, Russell Elliott\textsuperscript{1}, Harikrishna S. Kuttivelil\textsuperscript{1}, Katia Obraczka\textsuperscript{1}} 
     
    \IEEEauthorblockA{\textsuperscript{1}University of California, Santa Cruz \\Santa Cruz, California, USA \\ \{nbhatia3, rdelliot, hkuttive, obraczka\}@ucsc.edu}
    \IEEEauthorblockA{\textsuperscript{2}Independent, Boston, Massachusetts, USA \\ pranayko021@gmail.com} 
}

\maketitle

\begin{abstract}
Indoor Wi-Fi positioning remains a challenging problem due to the high sensitivity of radio signals to environmental dynamics, channel propagation characteristics, and hardware heterogeneity. Conventional fingerprinting and model-based approaches typically require labor-intensive calibration and suffer rapid performance degradation when devices, channel or deployment conditions change.
In this paper, we introduce Locaris, a decoder-only large language model (LLM) for indoor localization. Locaris treats each access point (AP) measurement as a token, enabling the ingestion of raw Wi-Fi telemetry without pre-processing. By fine-tuning its LLM on different Wi-Fi datasets, Locaris learns a lightweight and generalizable mapping from raw signals directly to device location.
Our experimental study comparing Locaris with state-of-the-art methods consistently shows that Locaris matches or surpasses existing techniques for various types of telemetry. Our results demonstrate that compact LLMs can serve as calibration-free regression models for indoor localization, offering scalable and robust cross-environment performance in heterogeneous Wi-Fi deployments. Few-shot adaptation experiments, using only a handful of calibration points per device, further show that Locaris maintains high accuracy when applied to previously unseen devices and deployment scenarios. This yields sub-meter accuracy with just a few hundred samples, robust performance under missing APs and supports any and all available telemetry.
Our findings highlight the practical viability of Locaris for indoor positioning in the real-world scenarios, particularly in large-scale deployments where extensive calibration is infeasible.

\end{abstract}


\section{Introduction}

Indoor positioning is critical in airports, hospitals, malls, retail, smart buildings, and for home- and elder care applications~\cite{9531633,farahsari2022survey,correa2017review}. However, most existing Wi-Fi telemetry-based localization systems are trained in static, controlled environments and need to be re-trained to adjust to new scenarios. 
Additionally, their accuracy degrades under everyday dynamics such as moving objects and new walls. Since Wi‑Fi telemetry is vendor‑specific, changing Wi‑Fi hardware~\cite{dwiyasa2016survey,deak2012survey} also poses challenges. 

Another important consideration is that there is a wide range of Wi-Fi telemetry types, each offering different trade-offs between availability and accuracy. Notably, RSSI which is widely available across a variety of devices, provides limited precision. On the other hand, less ubiquitous telemetry such as Channel State Information (CSI) allows highly accurate localization, yet is  rarely available due to specialized hardware and firmware requirements. This imbalance highlights the need for localization systems that can adapt to and operate in varying levels of telemetry availability; leveraging whichever signals are present while still achieving robust and accurate performance.
Additionally, Wi-Fi telemetry is unstructured and non‑standardized - this means that traditional ML models, which rely on fixed‑length vectors, suffer from padding, feature sparsity, and loss of information when the underlying Wi-Fi infrastructure changes (e.g., AP settings are modified). Consequently, a model trained in one environment fails to generalize to others, resulting in costly, time-consuming re‑engineering for each new deployment~\cite{twala2009empirical,nessa2020survey,roy2022survey,10609490,8692423,cominelli2023exposing}.

Decoder-only transformers, especially Large Language Models (LLMs), are a natural fit for Wi-Fi-telemetry based indoor localization. They offer five key advantages: 
\begin{itemize} 
\item \textbf{Schema‑free, variable‑length ingestion:} Each telemetry reading can be treated as a token. As such, handling variable number of APs and mixed telemetry modalities can be achieved without padding, aggregation, or imputation.
\item \textbf{Few‑shot learning:} Unlike traditional methods that struggle as dimensions increase (curse of dimensionality~\cite{beyer1999nearest}), LLMs exploit this large space to encode rich patterns and relationships. This property makes them especially powerful for few-shot learning, where only a handful of examples are needed to generalize across tasks even in domains like localization that were not part of their original training~\cite{hegselmann2023tabllm}. 
\item \textbf{Wireless signal propagation aware attention:} Self‑attention captures cross‑token structure created by multipath propagation, Non‑Line‑of‑Sight (NLOS) conditions, and interference, enabling holistic learning of long‑range dependencies that better fit RF phenomena than manual feature engineering. 
\item \textbf{Temporal pattern learning:} The autoregressive structure and positional encodings of LLMs capture subtle spatio-temporal variations in RF signals that arise from even small positional changes and environmental dynamics.
\item \textbf{Cross‑environment generalization:} LLMs generalize well across domains, allowing one model to operate across environments, vendors, and hardware, cutting deployment costs and accelerating rollouts.~\cite{pires2019multilingual,google2024dolphingemma,liu2025can,10.1145/3677846.3677854} 
\end{itemize}

We introduce \textbf{Locaris} (Latin for "you are located"), a compact, decoder-only language model specifically designed for indoor localization using raw Wi-Fi telemetry. Our system treats each measurement, e.g., from Access Points, as a token, allowing the model to process unstructured, variable-length, multi-modality telemetry without manual feature engineering across vendors. Importantly, for resource-constrained computing deployments, Locaris employs a parameter-efficient adaptation strategy through weight freezing and adaption through Low-Rank Adaptation (LoRA) modules~\cite{hu2021loralowrankadaptationlarge} on attention projections. We further assess different approaches to shrink memory usage while preserving localization accuracy. Additionally, the architecture utilizes deterministic regression outputs through an additional layer, giving consistent deterministic output for real-world use-case.

Our experimental evaluation of Locaris demonstrates that it can achieve state-of-the-art performance across multiple telemetry types and vendor configurations, eliminating vendor-specific calibration requirements while supporting rapid adaptation to new environments with minimal data requirements. 

The system maintains robust performance even with dropped APs or missing modalities, addressing key problem for real-world deployments where network conditions can be unpredictable.

In this paper, we make the following key contributions:
\begin{itemize}
    \item \textbf{Novel LLM-MLP hybrid architecture for Wi-Fi Localization:} To our knowledge, Locaris is the first system to repurpose decoder-only LLMs for indoor localization, demonstrating they are a powerful feature extractor when paired with lightweight MLP regression head.
    \item \textbf{Parameter-efficient adaptation for cross-environment deployment:} We develop a adaptation strategy using LoRA modules exclusively on attention projections while keeping the entire LLaMA backbone frozen, enabling rapid fine-tuning.
    \item \textbf{Universal, telemetry-agnostic architecture:} We introduce a generalized framework that handles variable-length multi-modality telemetry (e.g., RSSI, FTM) from diverse vendor configurations, producing deterministic regression outputs for real-time applications. 
    \item \textbf{Raw telemetry processing with graceful degradation:} We demonstrate that LLMs can learn directly from unstructured telemetry data without manual preprocessing, while gracefully degrading under incomplete telemetry or access points.
\end{itemize}

The remainder of this paper is organized as follows. Section~\ref{sec:background} provides background on Wi-Fi telemetry and reviews related work on traditional localization approaches and LLM applications. Section~III details Locaris' architecture and methodology. Section~IV describes our current implementation of Locaris and the experimental setup we used to evaluate Locaris' perfomance. Section~V presents our evaluation results and Section~VI contains discussion. Finally, Section~VII concludes the paper with implications for practical Wi-Fi localization deployment.

\section{BACKGROUND AND RELATED WORK}\label{sec:background} 

We start with a brief overview of Wi-Fi telemetry and then describe state-of-the-art localization approaches.

\subsection{Wi-Fi Telemetry}
Wi-Fi telemetry data provides various metrics, with each offering trade-offs between accuracy, complexity, and practicality for indoor positioning. RSSI (Received Signal Strength Indicator) is the most widely available method but offers low accuracy. Time-based approaches, such as Fine Time Measurement (FTM), provide higher accuracy with moderate availability. Channel State Information (CSI) delivers the highest accuracy by capturing fine-grained channel characteristics, though it requires specialized hardware and firmware, making it less available. These telemetry sources can also be fused with other sensors like gyroscopes, magnetometers, camera or Bluetooth to further enhance performance~\cite{MSIndoorLoc2021,lymberopoulos2017microsoft,roy2022survey,8692423}.

\noindent
{\bf RSSI:}
The Received Signal Strength Indicator (RSSI) is a metric used to quantify the strength of a received signal during wireless communication~\cite{pagano2015indoor}. 
RSSI-based techniques are commonly used because they are easily accessible and involve minimal overhead~\cite{mistry2015rssi}. RSSI values are often scaled based on the specific hardware and manufacturer specifications, meaning there is no standardized scale to compare RSSI values between devices from different manufacturers. Fingerprinting approaches~\cite{yiu2017wireless} are common, however, they are sensitive to environmental noise and device heterogeneity. Hence, even after enhancement of ML and sensor fusion~\cite{singh2021machine}, RSSI-based approach often shows high localization error. To reduce the error, RSSI is used in conjunction with other telemetry data through \textit{sensor fusion}, the process of combining telemetry data from different sensors. 

\noindent
{\bf FTM:}
IEEE 802.11-2016~\cite{9733026} included the first generation of the FTM protocol that sends a burst of frames and averages the round-trip-time (RTT).  FTM can perform significantly better than RSSI~\cite{ibrahim2018verification}. However, not all commercial devices support it; accuracy is affected by clock skew and drift and needs adequate calibration. Its performance suffers in Non-Line-Of-Sight environments (NLOS), where positioning accuracy deteriorates and errors can be as high as 5-10 meters~\cite{bullmann2020comparison,ibrahim2018verification}.

\noindent
\subsection{LLMs and Existing Applications}
Transformer models have demonstrated remarkable adaptability across diverse domains. In particular, as shown in ~\cite{pires2019multilingual}, large language models (LLMs), which can be realized by generative pre-trained transformers (GPTs), are trained primarily in English, but still perform well in other languages. Since then, transformers' capacity for generalization has expanded far beyond natural language processing to include applications such as computer vision, robotics, structured code generation, modeling vital signs, e.g., electrocardiogram (ECG) signals, and even interpreting dolphin vocalizations~\cite{google2024dolphingemma,liu2025can,10.1145/3677846.3677854}. 

Transformers can be broadly classified into two types -- those that can handle all tokens at once (encoders) and those that can only handle past tokens (decoders)~\cite{qorib2024decoder}.
Encoder-based transformers such as BERT~\cite{koroteev2021bert} take full input and process it simultaneously. This requires access to the whole input sequence before any processing can begin (non-causal). However, they cannot be used when data is available in real time (such as Wi-Fi packets) and are more suitable for offline tasks such as recovering lost network packets~\cite{zhao2024mininglimiteddatasufficiently} or data imputation~\cite{cesar2023bert}. 
Decoder-based transformers generate an output sequence that is not dependent on the future element (causal). Decoder-only transformers are also referred to as \textit{autoregressive}, in that each output is generated one step at a time based on current and past elements; thus decoder-only architectures are well-suited for real-time applications. Contemporary LLM systems, such as OpenAI's GPT~\cite{ai2023gpt} and Meta's LLaMA~\cite{touvron2023llama}, adopt a decoder-only architecture. Throughout the paper, we use the term Generative Pretrained Transformers (GPT) for decoder-based transformers.

\section{Locaris}\label{sec:locaris}
\begin{figure}[htbp]
    \centering
    \includegraphics[width=\linewidth]{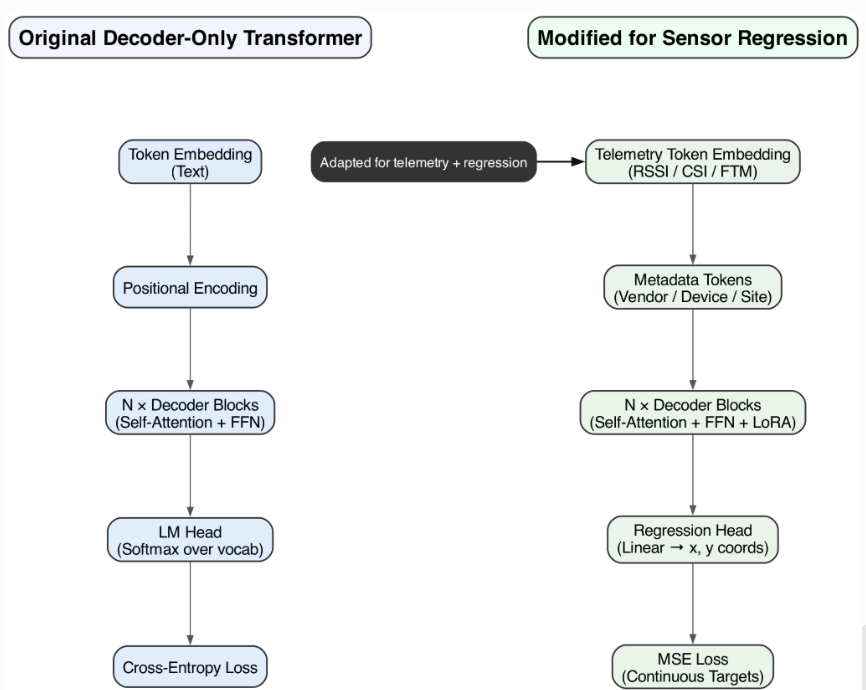}
    \caption{Traditional- \& Locaris' GPT.}
    \label{fig:locaris-overview}
\end{figure}

\begin{figure*}[ht!]
    \centering
    \includegraphics[width=\textwidth]{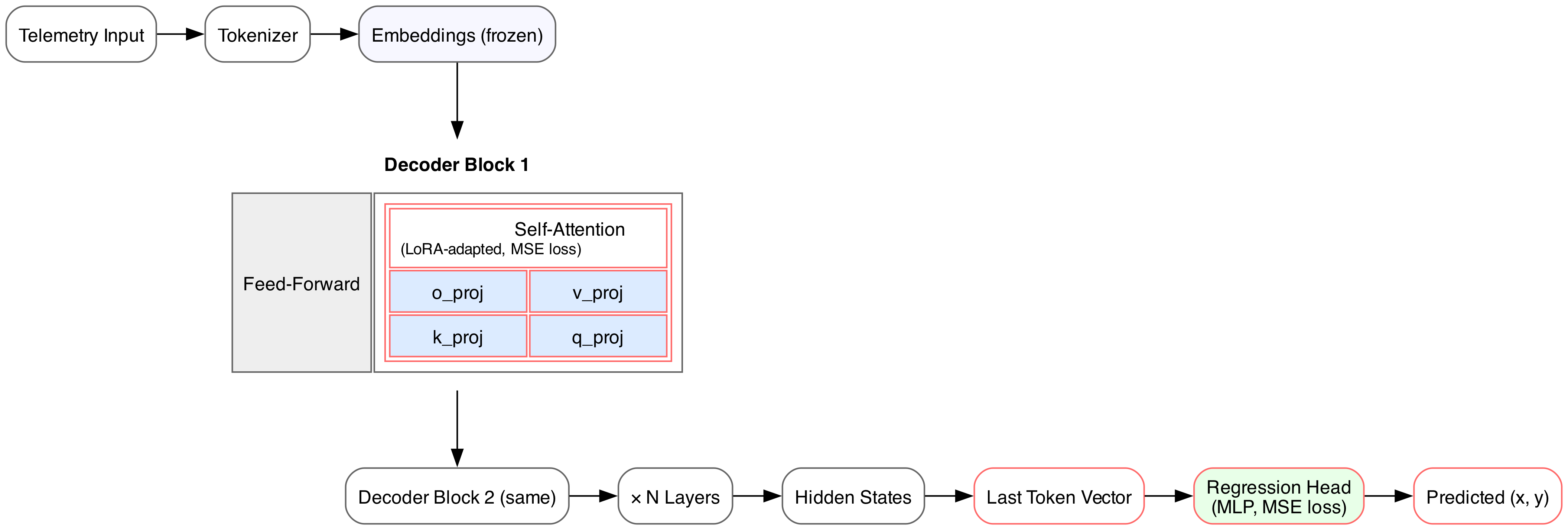}
    \caption{Locaris: System Architecture}
    \label{fig:llm-flow}
\end{figure*}

In this section, we describe Locaris’ decoder-only indoor localization system which addresses the core challenges of indoor Wi-Fi localization: device heterogeneity, variable infrastructure, and deployment stability. This is achieved through a fundamentally different approach by treating localization as a language problem. Rather than engineering fixed-size feature vectors from Wi-Fi measurements, Locaris processes raw telemetry as sequences of tokens, similar to how language models process text. As illustrated in Figure 1, the system consists of three main components: (1) a token-based input representation that handles variable-length, multi-modal telemetry without preprocessing, (2) a frozen pre-trained language model backbone with lightweight adaptation layers, and (3) a simple regression head that maps learned representations to spatial coordinates. This architecture allows the model to generalize across environments, vendors, and deployment configurations while maintaining parameter efficiency. The key hypothesis is that Wi-Fi measurements from multiple access points can form structured sequences with inherent relationships, similar to words in a sentence, and thus signal propagation patterns, multipath effects, and spatial dependencies can be captured through self-attention mechanisms originally designed for language.

In the remainder of this section, we describe each component of Locaris' architecture, as illustrated in Figure 2, in detail

\subsection{Telemetry Input}
Locaris represents each Wi-Fi measurement as a textual token sequence. For a device receiving signals from multiple access points (APs), telemetry input takes the form:
\begin{verbatim}
"AP1 RTT: <v> AP2 RTT: <v> ... AP5 RTT: <v>
 AP1 RSS: <v> AP2 RSS: <v> ... AP5 RSS: <v>"
\end{verbatim}  
This representation offers multiple benefits over traditional fixed-size vectors. First, it naturally handles variable numbers of access points. For example, if an AP is unavailable, it is simply omitted from the sequence rather than requiring imputation (due to fixed length) or masking. Second, different telemetry modalities (RSSI, FTM, sensor and PHY data, vendor info, device data etc) can be mixed freely without defining rigid schemas. Third, the sequential structure preserves relationships between measurements that would be lost in flattened feature vectors.

\subsection{Transformer Backbone}
At the center of Locaris is a pre-trained decoder-only transformer. While we base our current Locaris implementation on LLaMA‑3.2-1B, our approach generalizes to any decoder‑based large language model system. Rather than training from scratch, we use the model's existing ability for sequential reasoning and pattern recognition. The base model remains unchanged, i.e., all original weights are kept constant, which provides two key benefits: it keeps the learned attention patterns and positional encodings that naturally suit sequential wireless data, and it substantially reduces computational requirements.
Adaptation to the localization task occurs through Low-Rank Adaptation (LoRA) modules~\cite{hu2022lora,hu2021loralowrankadaptationlarge}, which are compact trainable matrices inserted into the attention mechanism of each transformer layer. These adapters modify how the model attends to different parts of the input sequence, allowing it to learn localization-specific patterns (e.g., that nearby APs should have higher attention weights) without disrupting the underlying sequence modeling capabilities.

\subsection{Sequence Aggregation and Prediction}
The output of the Locaris' transformer layer pipeline is 
a fixed-size representation 
corresponding to the last token in the sequence. This is a vector containing information from all access points and measurements through the transformer's attention mechanism.
A lightweight two-layer multilayer perceptron (MLP) stage then maps this representation to 2D (or 3D) spatial coordinates. Unlike traditional language models that generate probabilistic tokens, Locaris incorporates an MLP-based deterministic regression head, which ensures its outputs are stable and continuous real-valued position estimates.

\subsection{Training Overview}
The model tries to minimize mean squared error (MSE) between predicted and ground-truth coordinates. This differs from standard language model training in two key ways: (1) there is no next-token prediction or cross-entropy loss, and (2) we supervise only the final coordinate output rather than intermediate sequence positions. Training updates only LoRA adapter weights and regression head parameters, which represent less than 1\% of the full model. This design enables training on single GPUs with limited memory capabilities and allows rapid adaptation to new deployment sites with minimal data during training.
\subsection{Inference and Computation Cost}
At inference time, Locaris performs a single forward pass: the tokenized measurement sequence passes through the transformer, the last-token representation is extracted, and the regression head outputs coordinates. The computational complexity is dominated by self-attention operations, however, Wi-Fi telemetry sequences are typically short (10-50 tokens depending on the number of APs and modalities), resulting in millisecond-scale latency on commodity GPUs or seconds on mobile CPUs~\cite{hoffmann2022trainingcomputeoptimallargelanguage}.

\section{System Evaluation Methodology}

\subsection{Datasets}

\begin{figure*}[htbp]
    \centering
    \includegraphics[width=\textwidth]{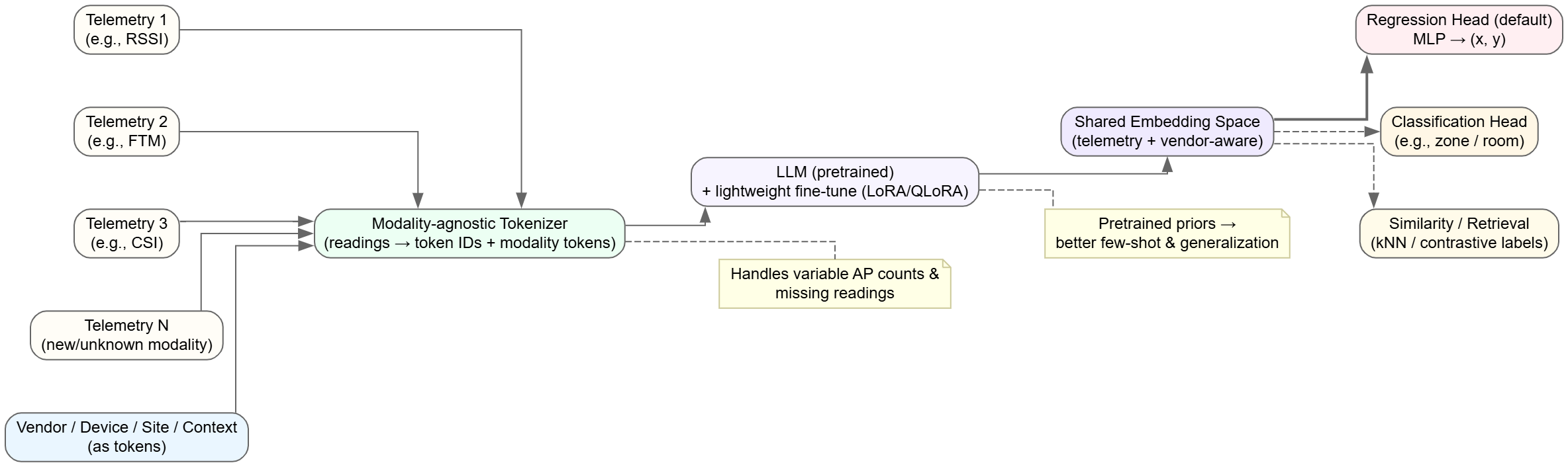}
    \caption{Telemetry pipeline}
    \label{fig:wifigpt-overview}
\end{figure*}

We demonstrate the accuracy of our system using two datasets.
The first is the SODIndoorLoc dataset, a large-scale Wi-Fi fingerprinting benchmark covering three buildings, multiple floors and rooms, with data collected from different users and devices. It provides both dense and averaged RSSI fingerprints along with ground-truth coordinates, making it suitable for comprehensive localization benchmark and cross-vendor testing.
The second data set is publicly available~\cite{zenedo}, combining Wi-Fi fine-time measurement (FTM) and received signal strength indicator (RSSI) data collected in three environments: a lecture theater (LOS), an office (mixed LOS-NLOS), and a corridor (NLOS) to improve positioning accuracy under varying conditions. Using these two datasets underscores our model's adaptability and robustness across telemetry and environmental scenarios.
\subsubsection{SODIndoorLoc}
\begin{table}[htbp]
\centering
\begin{tabular}{l l l l}
\hline
Feature & CETC331 & HCXY & SYL \\
\hline
Train samples & 955   & 11370 & 8880 \\
Test samples  & 840   & 860   & 1020 \\
Floors        & 3     & 1     & 1    \\
Spacing       & $\sim$1.2 m & $\sim$1.2 m & $\sim$0.5 m \\
\hline
\end{tabular}
\caption{Dataset characteristics of CETC331, HCXY, and SYL \cite{bi2022supplementary}.}
\label{tab:data_features}
\end{table}

We use the SODIndoorLoc dataset, which consists of Wi-Fi fingerprint measurements collected in three buildings: CETC331, HCXY, and SYL. Each building includes detailed information on pre-installed access point (AP) locations and specifications, along with corresponding CAD drawings. Sampling points are arranged at regular intervals, with approximately 1.2 meters between points in CETC331, and HCXY, and 0.5 meters in SYL. Each sample records RSSI values from each AP, ranging from -104 dBm to 0 dBm. A value of 100 indicates missing RSSI measurements. Every sample also includes metadata such as spatial coordinates, floor ID, building ID, user ID, phone ID, and sample count. All tests were conducted using three distinct random seeds.
\subsubsection{Wi-Fi RSS and RTT dataset with different LOS conditions for indoor positioning}
\begin{table}[ht]
\centering
\caption{Wi-Fi RSS \& RTT dataset with different LOS conditions for indoor positioning\cite{feng2024wifi}}

\begin{tabular}{|l|c|c|c|}
\hline
\textbf{Data features} & \textbf{Lecture Theatre} & \textbf{Office} & \textbf{Corridor} \\
\hline
Testbed (m$^2$) & 15 $\times$ 14.5 & 18 $\times$ 5.5 & 35 $\times$ 6 \\
Grid size (m$^2$) & 0.6 $\times$ 0.6 & 0.6 $\times$ 0.6 & 0.6 $\times$ 0.6 \\
Number of RPs & 120 & 108 & 114 \\
All samples & 7,200 & 6,480 & 6,840 \\
Training samples & 5,280 & 4,860 & 5,110 \\
Testing samples & 1,920 & 1,620 & 1,740 \\
Wi-Fi condition & LOS & LOS-NLOS & NLOS \\
\hline
\end{tabular}
\end{table}

We use a publicly available Wi-Fi dataset that comprises various LOS and NLOS environments~\cite{zenedo}. It contains an extensive selection of samples from multiple reference points (RP) in three different scenarios: lecture theater (LOS), corridor (mixed LOS / NLOS) and office (NLOS) (Table II). Data are collected using a Google AC-1304 Wi-Fi router and Pixel 3 smartphones. The data set contains RTT and RSSI measurements, as both the Access Point (AP) and the Station (STA) support the FTM protocol. All tests were conducted using five distinct random seeds.

\subsection{Experiments Overview}
\subsubsection{SODIndoor Dataset}
For our experiments, we used data from the CETC331, HCXY, and SYL buildings. Rather than constraining the input to a fixed set of access points, we constructed the input as a flexible text prompt by appending valid RSSI values together with vendor identifiers, scene descriptions, and building labels. This allowed the model to interpret heterogeneous information in natural language form and learn associations useful for localisation. Each building followed the default train/test split specified by \cite{bi2022supplementary}, and the models were trained to regress directly to the X and Y coordinates without additional feature standardization and averaged over 3 seeds. To contextualize performance, we also report comparisons against baseline regression methods following the evaluation protocol of \cite{bi2022supplementary}, as summarized in Table~\ref{tab:data_features}.

\subsubsection{FTM RSSI Dataset}
\begin{itemize}
    \item \textbf{Cross-Environment Transfer:} Here, we investigate its cross-environment learning and generalization through few-shot learning experiments. We compare the 1B parameter model against CNN, KNN, MLP, LGBM, and simple transformer baselines. These models were selected based on their top performances in Microsoft's 2021 Indoor Localization Competition \cite{MSIndoorLoc2021}. All non-linear baselines underwent hyperparameter optimization through 30 Optuna~\cite{Optuna} trials to ensure fair comparison. We also included an additional bit to indicate the environment for the baseline models. Our evaluation method tests true cross-environment transfer: models are trained on two source environments and then provided a fraction (1-100\%) of the training data from a held out target environment before evaluation on that environment's test set. This setup simulates real-world deployment where pre-trained models must adapt to new buildings and spaces with minimal site-specific calibrations. 
    \item \textbf{Telemetry Ablation:} We evaluate the effect of different input modalities as a demonstration of the model’s ability to handle flexible inputs. For Locaris, input samples are formed by directly concatenating the available telemetry from all access points. In the full setting, both FTM and RSSI values are included; in the ablation settings, we simply provide whichever modality is available (FTM-only or RSSI-only) without requiring placeholders. In contrast, the baseline models do not support variable-length or modality-flexible inputs, and therefore rely on masked values to represent missing telemetry (e.g., assigning extreme placeholders such as –200 dBm for RSSI or 100000 ns for FTM). This distinction allows Locaris to handle heterogeneous and incomplete telemetry natively, while baselines depend on custom masking schemes. The data set includes paired FTM (ns) and RSSI (dBm) values for each AP (five APs in most environments, four in corridor's case). 
    \item \textbf{Access Point (AP) Ablation:} To evaluate the performance of the model under limited connectivity, we conducted an ablation study of the access point (AP) with five available APs. For the 1-AP dropout case, we remove one AP at a time in both the single environment (model trained \& tested on one environment) and all environment (model trained on all three environments and tested on all environments) settings. The dropout is applied uniformly in a circular pattern and the results are averaged across rotations. For the 2-AP dropout case, evaluation is limited to the Single Environment setting due to space constraints. All unique AP pairs are tested, though we skip the Corridor environment since it only has four APs and trilateration requires at least three.
\end{itemize}

\subsubsection{Performance Metrics}

To evaluate Locaris’s performance, we utilize the following metrics:

\begin{itemize}
    \item \textbf{Mean Absolute Error (MAE)}: measures the average magnitude of prediction errors. MAE values are reported in meters (m), which reflect the average distance error between predicted and actual locations, a key factor in applications like asset tracking and indoor navigation.

    \item \textbf{Root Mean Squared Error (RMSE)}: measures the square root of the average squared differences between predicted and true coordinates. Since squaring emphasizes larger deviations, RMSE penalizes outliers more strongly than MAE. 

    \item \textbf{X Percentile Error}: represents worst-case performance in X\% of cases. This metric highlights Locaris’s robustness and reliability under challenging conditions.  A lower value indicates consistent accuracy even with noisy data, which is vital for dynamic environments.

    \item \textbf{Energy (samples per watt)}: measures the energy efficiency of the system, which is critical for battery-powered deployments. Higher values correspond to greater efficiency, meaning more samples can be processed per unit of energy.  

    \item \textbf{Throughput (samples per second)}: measures the processing speed of the system, reflecting how many samples can be handled in real time. This metric is essential for latency-sensitive applications such as indoor navigation and tracking. Higher values indicate faster inference and improved responsiveness in deployment.  

    \item \textbf{Memory (MB)}: represents the memory footprint of Locaris, a key consideration for resource-constrained devices. This value highlights the memory requirements of Locaris and its feasibility for deployment on embedded systems.

\end{itemize}

We evaluate the impact of quantization on both localization accuracy and system efficiency using the FTM-RSSI dataset. The results show how reducing precision affects error metrics (MAE, RMSE, P95) while also influencing throughput, memory footprint, inference time, and energy usage. This highlights the trade-off between maintaining high accuracy and achieving lightweight, efficient deployment on resource-constrained devices for various indoor localization applications.

\begin{table*}[htbp]
\centering
\caption{Average statistics of regression methods across CETC331, SYL, and HCXY datasets.}
\begin{tabular}{lccccc}
\hline
\textbf{Method} & \textbf{MAE (m)} & \textbf{RMSE (m)} & \textbf{50th (m)} & \textbf{75th (m)} & \textbf{95th (m)} \\
\hline
KNC     & 4.116  & 4.664  & 2.201  & 4.874  & 16.247 \\
KNR     & 3.758  & 4.178  & 2.703  & 4.768  & 10.293 \\
MLPC    & 18.183 & 17.483 & 12.301 & 25.269 & 54.864 \\
MLPR    & 8.022  & 6.384  & 6.202  & 11.444 & 19.168 \\
RFC     & 6.735  & 10.494 & 1.911  & 5.483  & 33.006 \\
\textbf{Locaris} & \textbf{2.840} & \textbf{6.140} & \textbf{3.880} & \textbf{7.540} & \textbf{10.280} \\
\hline
\end{tabular}
\end{table*}
\section{Results}

This section presents a comprehensive evaluation of Locaris's performance across various datasets and scenarios and a discussion of Locaris's ability to rapidly adapt to new environments.

\subsection{SODIndoor }

Table ~III reports the average performance of baseline regression methods compared to our proposed Locaris system across CETC331, SYL, and HCXY. Among traditional baselines, KNR and KNC achieve moderate accuracy, with mean absolute errors around 3–4 meters, though their performance deteriorates at higher percentiles (95th greater than 10 m). RFC shows lower median errors but suffers from large outliers, with 95th percentile errors exceeding 30 m. MLPC demonstrates instability, with both average and tail errors significantly worse than other baselines, indicating its limited suitability for RSS-based positioning.

In contrast, Locaris consistently outperforms all baselines. It achieves the lowest MAE (2.84 m) and RMSE (6.14 m) across datasets, representing an improvement of roughly 25–30\% over KNC and KNR. More importantly, Locaris significantly reduces high-error outliers, with a 95th percentile error of only 10.28 m, compared to 16.25 m for KNC and 33.01 m for RFC. 
\subsubsection{Take-aways}
This reduction in extreme errors highlights Locaris’s robustness and reliability, particularly in challenging real-world deployment scenarios where worst-case performance is critical. Unlike traditional baselines which require fixed preprocessing rules such as standardization of RSS values, selecting top-K access points, and enforcing dataset-specific thresholds, Locaris operates directly on in-range RSS values and encoded device metadata without such rigid constraints, offering a more flexible and generalizable approach across environments.

\subsection{FTM-RSSI}

\subsubsection{Cross-Environment Transfer}

\begin{figure}[htbp]
    \centering
    \includegraphics[width=\linewidth]{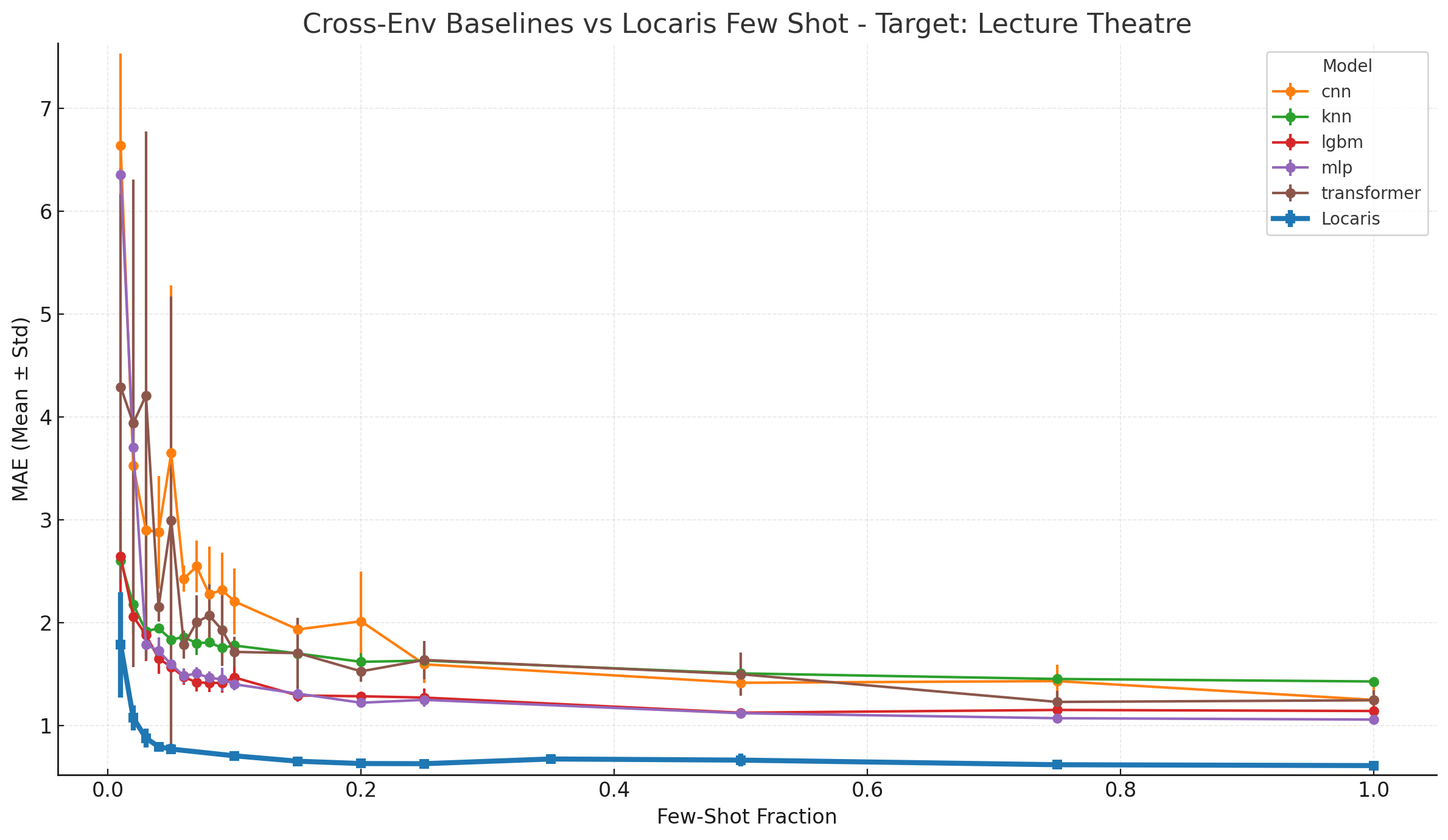}
    \caption{Locaris vs. baselines in few-shot learning (Lecture target).}
    \label{fig:lecture_few_shot}
\end{figure}

\begin{figure}[htbp]
    \centering
    \includegraphics[width=\linewidth]{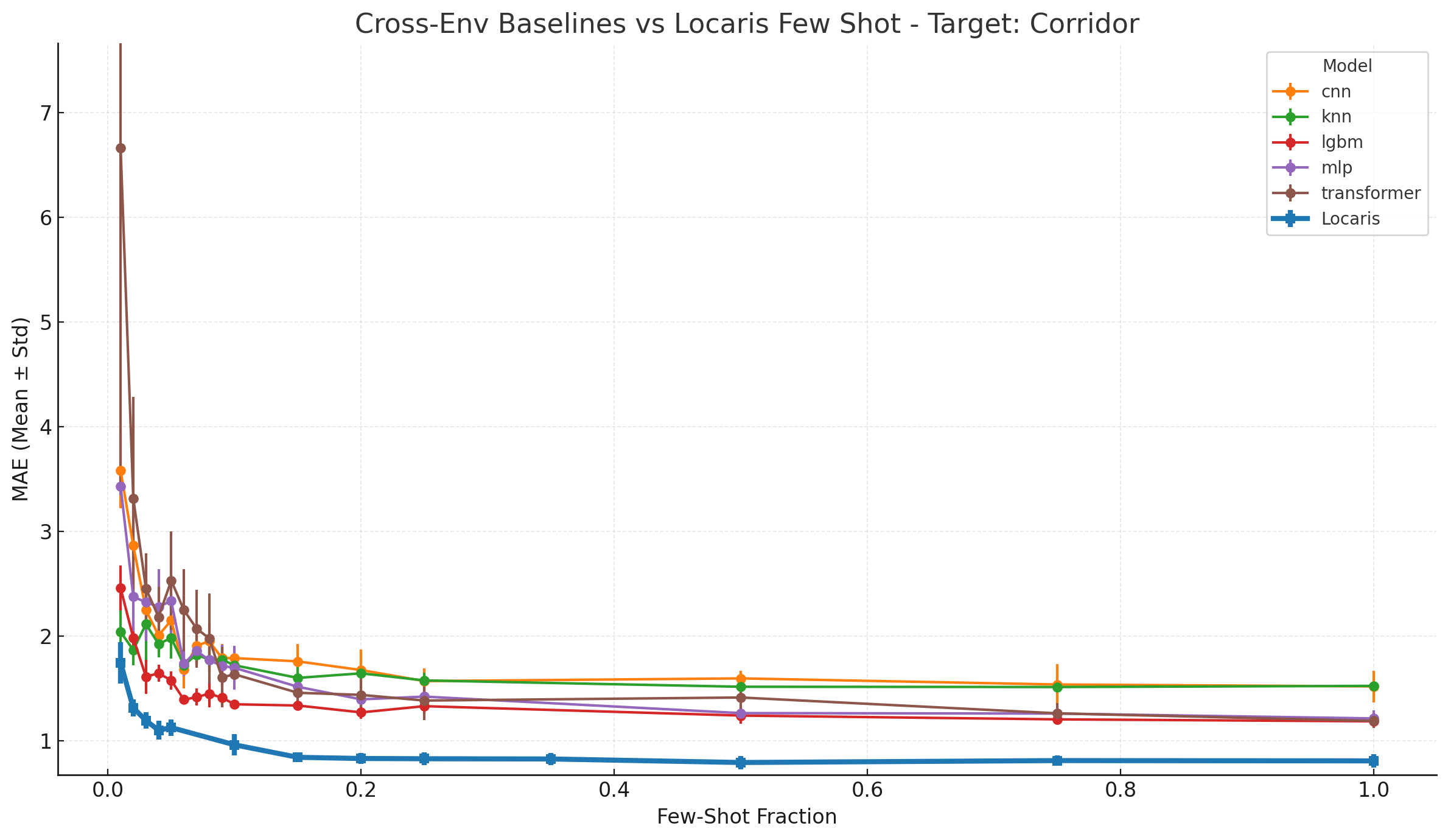}
    \caption{Locaris vs. baselines in few-shot learning (Corridor target).}
    \label{fig:corridor_few_shot}
\end{figure}

\begin{figure}[htbp]
    \centering
    \includegraphics[width=\linewidth]{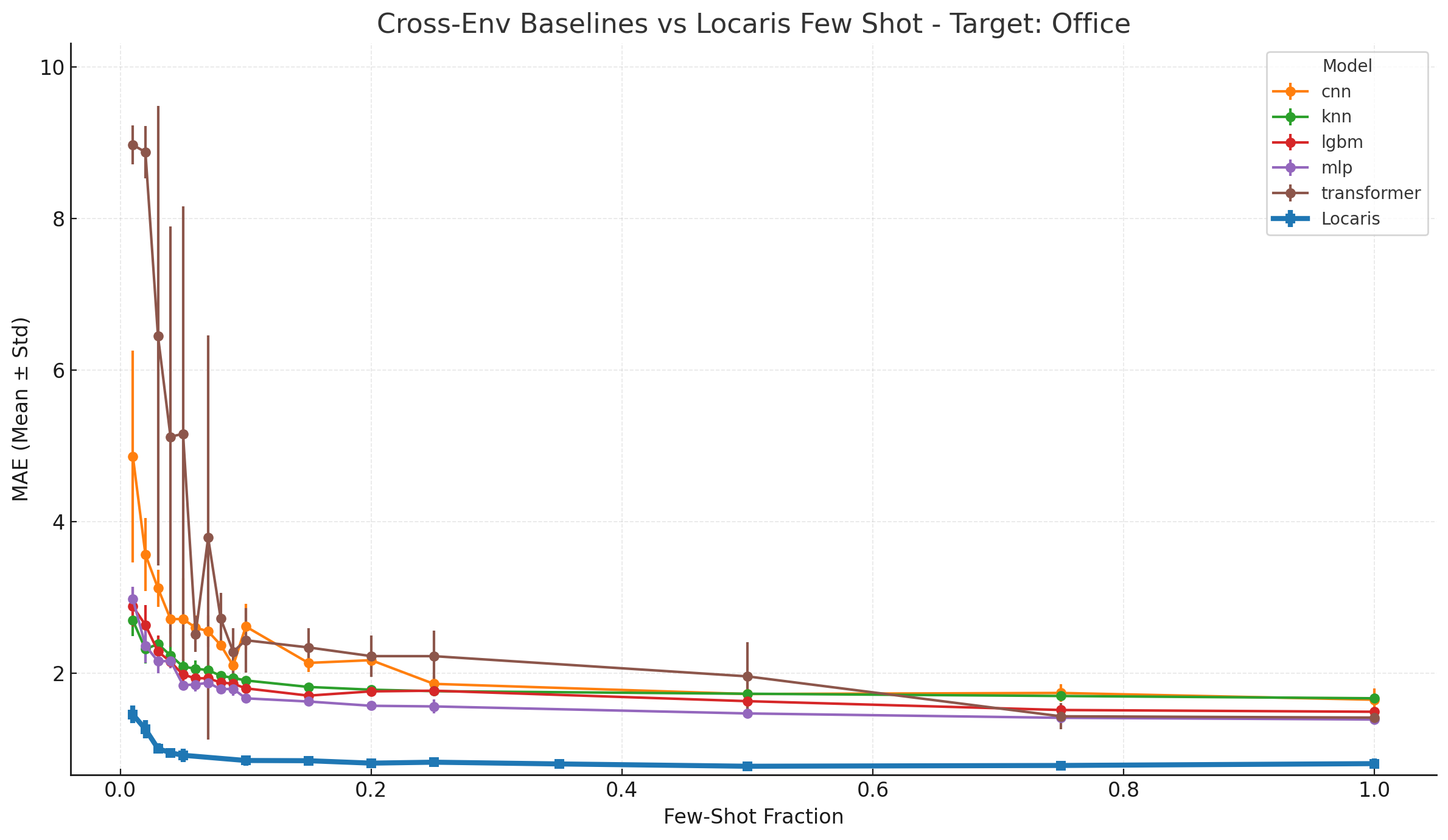}
    \caption{Locaris vs. baselines in few-shot learning (Office target).}
    \label{fig:office_few_shot}
\end{figure}

Across all cross-environment scenarios, Locaris achieves sub meter accuracy with significantly less target-specific data than baselines. The lecture theater target reaches the submeter performance the fastest, reaching 0.88 meter error with only 3\% of the target data, while the best baseline remains at 1.88m, a 53\% performance gap seen in Figure \ref{fig:lecture_few_shot}. The office target achieves 0.95m error in 4\% data, where the baselines lay at 2.15m error, a 56\% advantage for Locaris as seen in Figure \ref{fig:office_few_shot}. The corridor target requires the most adaption given its heavy multipath. It reaches a 0.96m error in 10\% of the data compared to 1.35 for the best baseline, a 29\% improvement as seen in Figure \ref{fig:corridor_few_shot}. Additionally, the baselines fail to achieve submeter accuracy even with 100\% of the training data in the lecture (1.14m), office (1.49m), and corridor (1.19m). Locaris's final performances reach 0.81m for corridor, 0.61, in lecture, and 0.8m in office. 

Across all target environments, Locaris reaches near optimal performance with 5-10\% of target training data, while baselines require 50-75\% to near comparable accuracy. This 10x reduction in calibration requirements is key for deployment.

\subsubsection{Telemetry Ablation}

\begin{figure}[htbp]
    \centering
    \includegraphics[width=1\columnwidth]{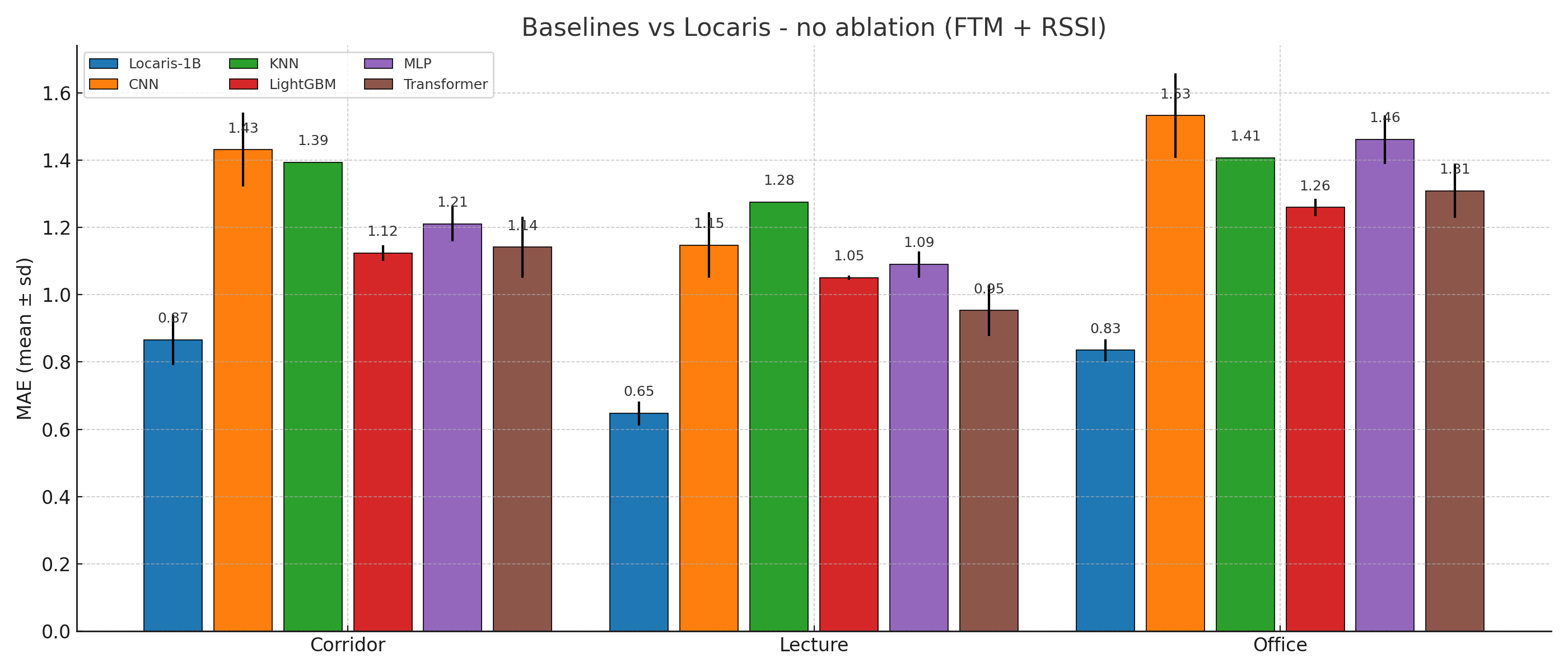}
    \caption{Combined FTM + RSSI (no ablation) results.}
    \label{fig:noablation-ftm-rssi}
\end{figure}
\begin{figure}[htbp]
    \centering
    \includegraphics[width=1\columnwidth]{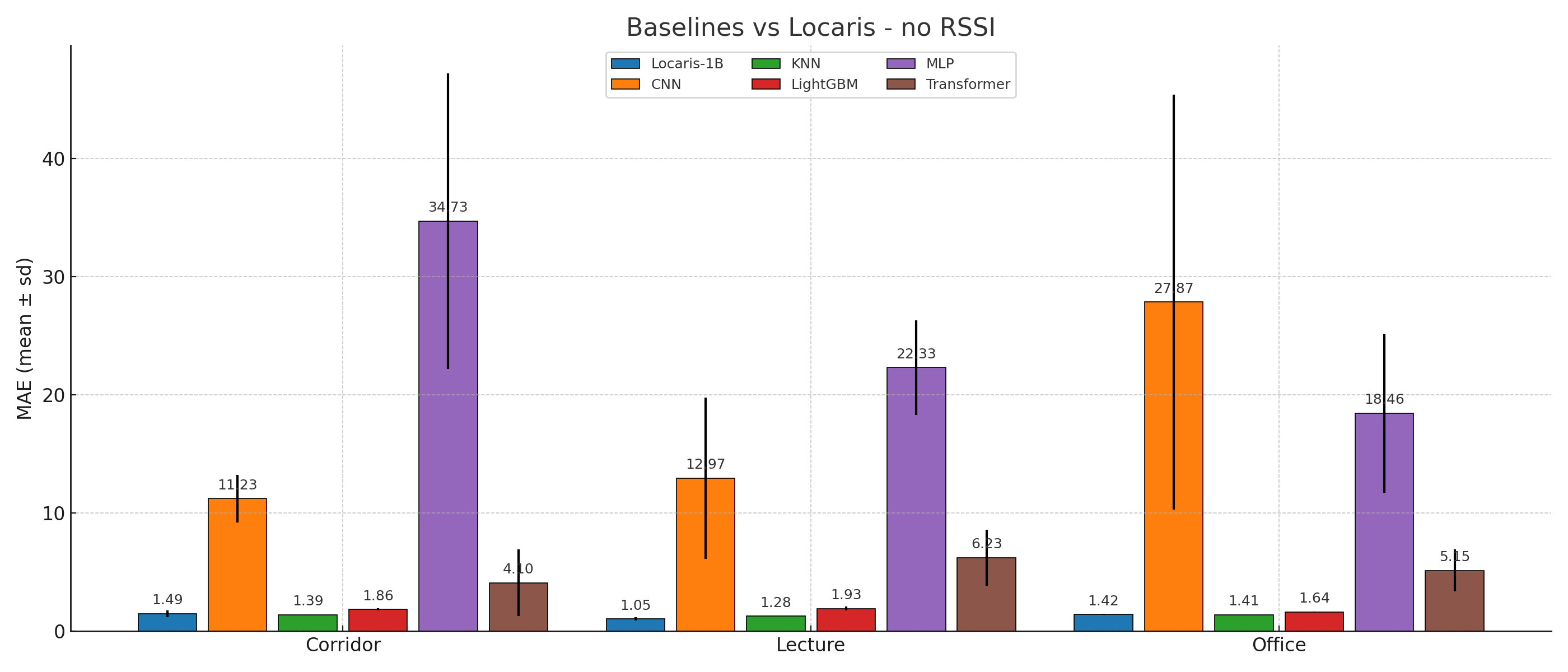}
    \caption{FTM-only ablation results.}
    \label{fig:ftm-only}
\end{figure}

\begin{figure}[htbp]
    \centering
    \includegraphics[width=1\columnwidth]{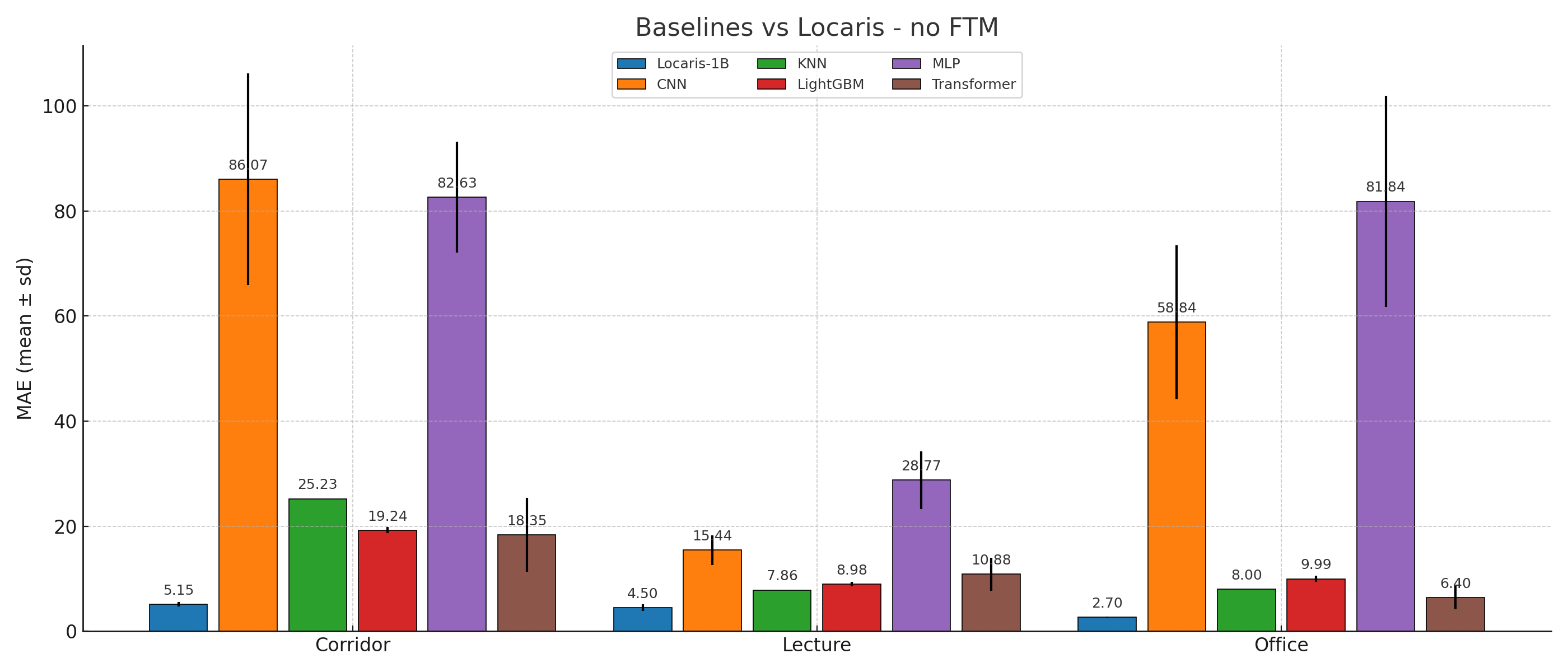}
    \caption{RSSI-only ablation results.}
    \label{fig:rssi-only}
\end{figure}
\begin{itemize}
    \item FTM+RSSI:
When both FTM and RSSI telemetry are available, Locaris consistently outperforms all baselines across all environments. In this full-input setting, it achieves sub-meter accuracy, with mean absolute errors (MAE) ranging from 0.65 to 0.87 meters, while the best baseline methods remain around 1.0–1.4 meters. CNN and kNN in particular perform worse than tree-based and transformer models, indicating that they struggle to effectively combine heterogeneous signals. These results highlight Locaris’s ability to jointly model FTM and RSSI, capturing nonlinear dependencies in wireless propagation that other methods fail to exploit.
    \item FTM-Only Ablation:  When only FTM telemetry is considered (no RSSI), most baselines degrade sharply, with CNN and MLP producing errors as high as 35 meters. In contrast, Locaris remains highly robust, maintaining MAEs between 1.05 and 1.49 meters. Interestingly, kNN achieves slightly better performance than Locaris in some instances (by approximately 6–7 cm), reflecting the advantage of direct distance-based heuristics when clean FTM data is available. Nevertheless, Locaris remains more consistent across environments and avoids the large variance exhibited by baseline models.
    \item RSSI-Only Ablation: 
The most challenging condition arises when FTM is removed, leaving only RSSI. In this setting, all baseline methods collapse, with CNN and MLP reaching errors between 60 and 100 meters. Locaris, however, demonstrates remarkable resilience, maintaining sub-5 meter accuracy across all environments, with MAE ranging from 2.86 to 4.71 meters. Locaris is able to extract meaningful spatial cues from weak telemetry signals where other models fail completely.

    \item Take-aways: 
Taken together, these results demonstrate that Locaris not only achieves state-of-the-art performance under ideal conditions, but also maintains robust accuracy under degraded or incomplete input modalities. While kNN is competitive in the FTM-only case, Locaris is the only model that generalizes well across all settings, avoiding catastrophic degradation when telemetry is missing. This robustness is particularly important for practical deployments, where access point capabilities vary and FTM support is often inconsistent. Locaris’s ability to deliver sub-meter localization in the full setting and sub-5 meter accuracy even under ablation makes it uniquely well-suited for real-world large-scale indoor localization systems.

\end{itemize}

\subsection{Access Point (AP) Ablation}
\begin{figure}[t]
  \centering
  \includegraphics[width=0.98\columnwidth,
                   height=0.28\textheight,
                   keepaspectratio,
                   trim=8 10 8 6,clip]{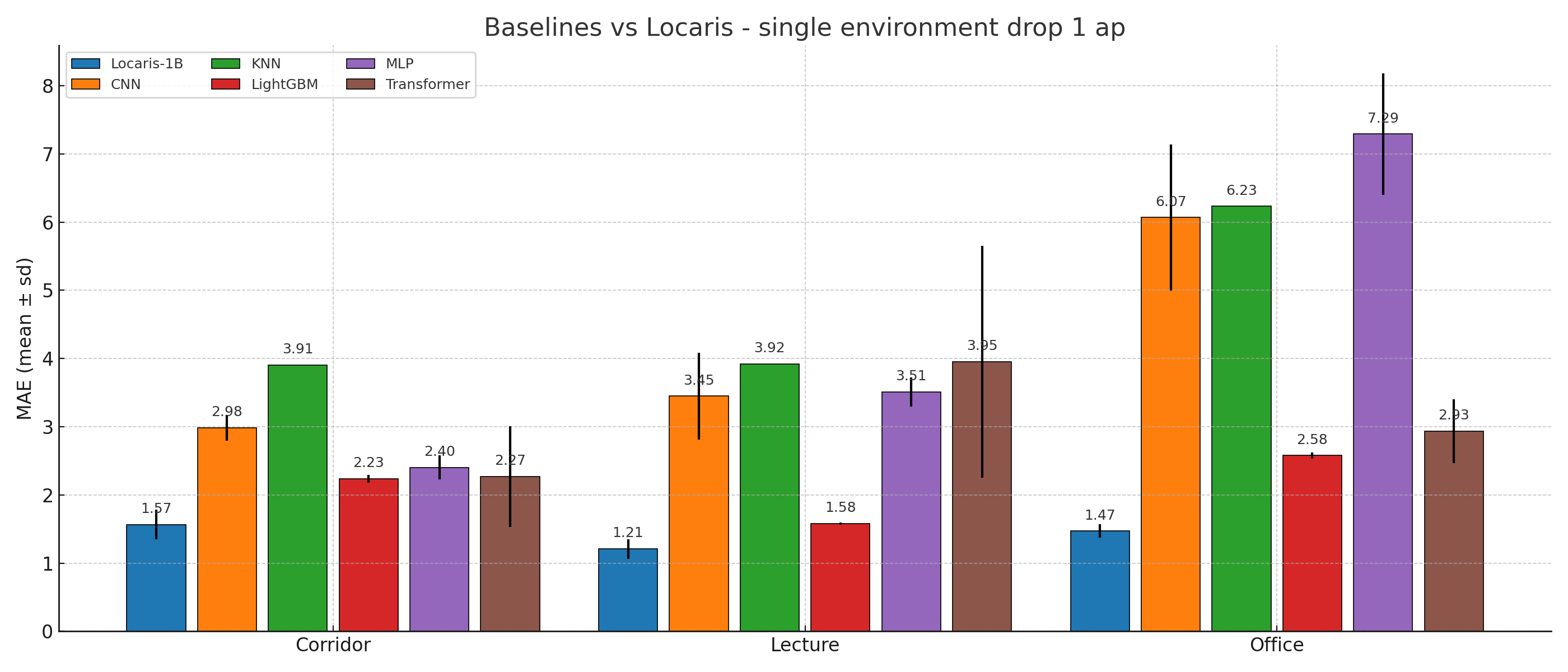}
  \caption{Single-environment results with a single AP removed.}
  \label{fig:singleap_singleenv}
\end{figure}

\begin{figure}[htbp]
    \centering
    \includegraphics[width=1\columnwidth]{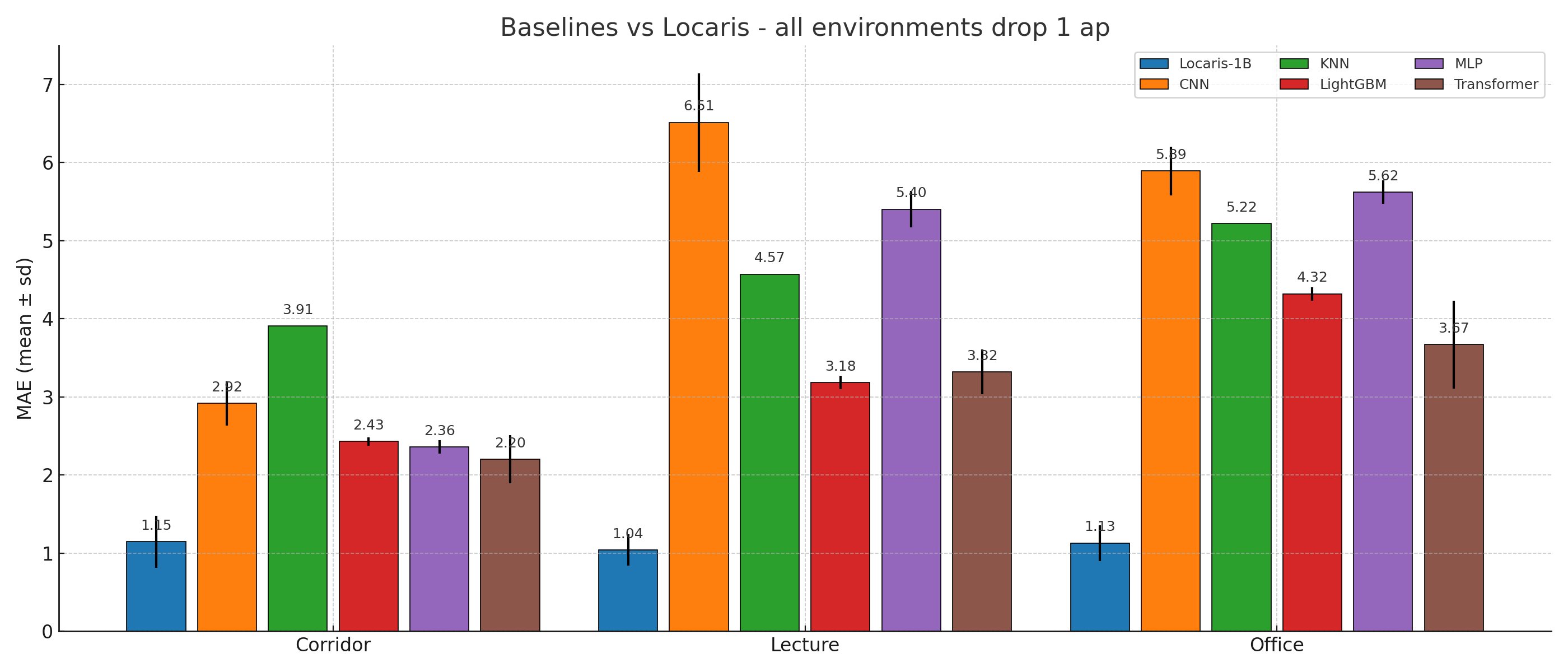}
    \caption{Ablation results across all environments with 1 AP dropped.}
    \label{fig:ablation-allenv-1ap-drop}
\end{figure}

\begin{figure}[t]
  \centering
  \includegraphics[width=0.98\columnwidth,
                   height=0.28\textheight,
                   keepaspectratio,
                   trim=8 10 8 6,clip]{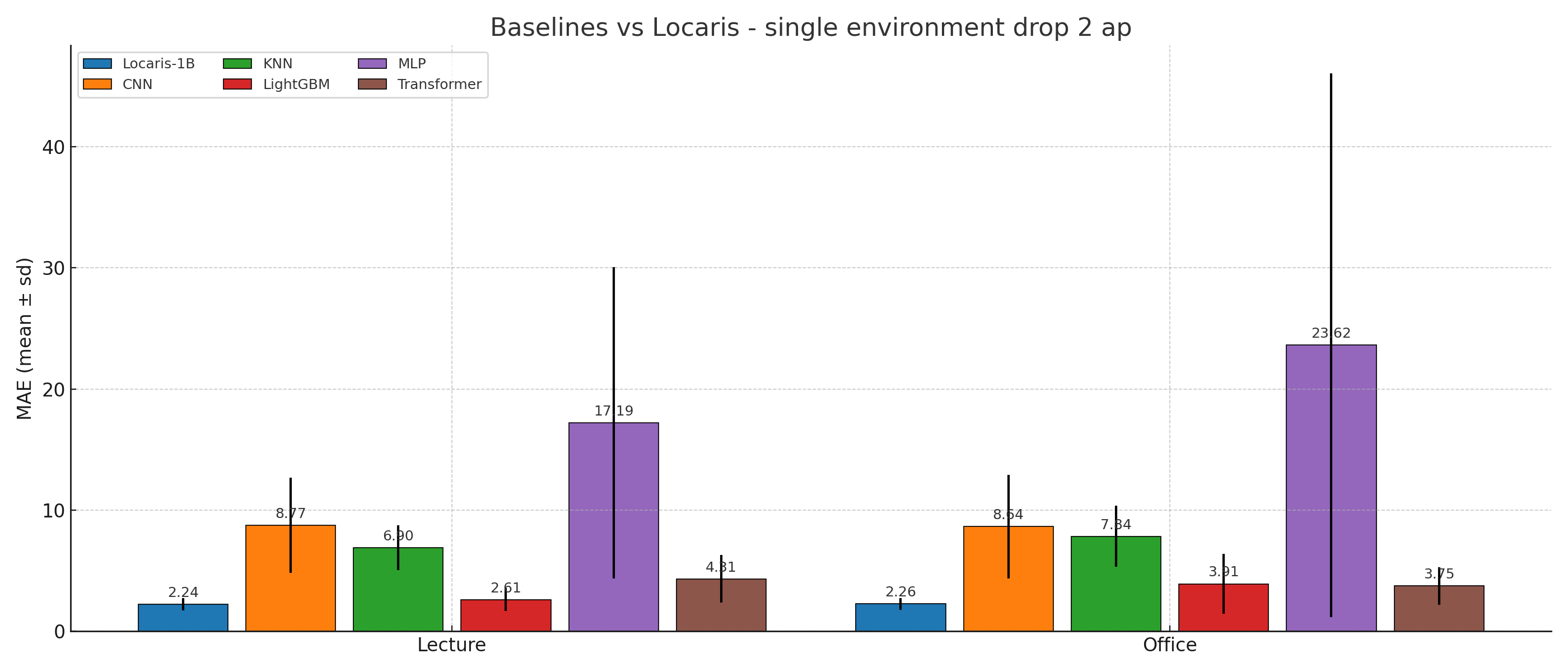}
  \caption{Single-environment results with 2 APs dropped — Lecture and Office.}
  \label{fig:singleenv_drop2ap_lecture_office}
\end{figure}
\subsubsection{Single Environment - 1 AP Drop}
Under the 1 AP drop condition in single-environment training, Locaris consistently achieves the lowest MAE across all scenes, ranging from 1.2–1.6 meters. This performance is substantially better than all baselines, which often exceed 3 meters in the corridor and 6 meters in office. CNN performs reasonably well in corridor (~3.0 m) but fails to generalize, with much higher errors in office. KNN and MLP both exhibit instability, with errors spiking in office. MLP in particular collapses with MAE above 7 meters, underscoring its poor robustness to infrastructure removal. LightGBM shows more stable behavior but remains consistently less accurate than Locaris. These results indicate that Locaris is uniquely resilient to access point loss, where all other methods degrade significantly.

\subsubsection{All Environments - 1 AP Drop}
When trained on all environments jointly with 1 AP dropped, Locaris demonstrates strong cross-environment generalization, maintaining MAE around 1.0–1.2 meters across corridor, lecture, and office. Notably, its performance improves slightly compared to single-environment training, suggesting that additional training data strengthens its robustness to missing infrastructure. By contrast, all baseline methods degrade compared to their single-environment performance. CNN and MLP struggle particularly in lecture and office, with errors exceeding 5–6 meters. KNN remains highly inconsistent, producing inflated errors in office, while LightGBM is more stable but still lags behind Locaris. These findings highlight Locaris’s scalability across diverse environments, showing that it benefits from broader training distributions while other models collapse.

\subsubsection{Single Environment - 2 AP Drops}
In the more challenging 2 AP drop scenario (lecture and office only, as corridor has only 4 APs), the gap between Locaris and baselines widens further. Locaris sustains low MAE of 2.5–3.5 meters, while CNN and KNN errors rise above 7–9 meters, and MLP collapses dramatically, reaching 18–26 meters depending on the environment. Transformer and LightGBM fare better but still trail Locaris by a significant margin, with errors in the 4–5 meter range. These results demonstrate that Locaris remains robust and stable even under heavy infrastructure degradation, while all baseline models suffer severe performance collapse.

\subsubsection{Take-aways}
The AP-drop experiments show that Locaris delivers higher stability and accuracy, and its performance degrades gracefully as the environmental parameter deteriorates. It consistently delivers the lowest MAE, scales effectively with cross-environment training, and maintains sub-4 meter performance even with two access points removed where baselines degrade catastrophically. This robustness strongly supports the case for Locaris as a practical, deployable solution in real-world localization systems, where access point availability is unpredictable.

\subsection{Accuracy and efficiency metrics for different quantization schemes}
\begin{table}[ht]
\centering
\caption{Accuracy and Efficiency Metrics for Different Quantization Schemes}
\resizebox{\columnwidth}{!}{%
\begin{tabular}{lccccccc}
\hline
\textbf{Type} & \textbf{MAE} & \textbf{RMSE} & \textbf{P95} & \textbf{Throughput} & \textbf{Memory} & \textbf{Time/Sample} & \textbf{Energy} \\
              & (m) & (m) & (m) & (samp/sec) & (MB) & (ms) & (samp/Wh) \\
\hline
FP16  & 0.7718±0.1000 & 1.0486±0.1656 & 2.7415±0.4335 & \textbf{25.85±0.97} & 6335±1309 & \textbf{38.74±1.28} & \textbf{772±27} \\
8-bit & \textbf{0.7880±0.1009} & \textbf{1.0697±0.1726} & \textbf{2.8093±0.4740} & 21.82±0.09 & 5407±1310 & 45.84±0.18 & 650±22 \\
4-bit & 0.9041±0.1535 & 1.2450±0.2957 & 3.2309±0.7819 & 23.92±0.88 & \textbf{4956±1311} & 41.85±1.40 & 698±87 \\
\hline
\end{tabular}%
}
\label{tab:quantization}
\end{table}


As shown in Table~\ref{tab:quantization}, quantization shows mixed outcomes on a Google Colab T4 GPU when evaluated across office, corridor, and lecture environments. While 8-bit and 4-bit models achieve substantial memory reductions (averaging 14.7\% and 21.8\% compared to FP16, respectively), their inference speed is consistently lower. This slowdown occurs because LoRA layers remain in FP16, requiring intermediate results to be converted back to FP16 for internal computation. 

\subsubsection{Take-aways}
Overall, quantization in this setting trades memory footprint for reduced energy efficiency and slower inference, with 8-bit offering a modest accuracy improvement at the cost of throughput, while 4-bit provides the best memory savings but significant accuracy degradation. These results indicate that FP16 remains preferable when latency and power efficiency are critical, whereas 4-bit may be suitable for memory-constrained deployments where accuracy loss is acceptable.

\section{Discussion} 

A central strength of Locaris lies in its robustness under degraded conditions. Unlike baselines, which often collapse under ablation, Locaris degrades gracefully, maintaining sub-4 meter accuracy. Its ability to significantly reduce extreme localization errors and learn quickly further reinforces its reliability, which is particularly important in real-world deployment scenarios where worst-case accuracy often dictates system usability.

Quantization presents a clear trade-off between memory footprint and inference efficiency. While reducing precision helps shrink model size, it comes at the cost of slower execution and higher energy consumption. The results show that 8-bit quantization offers modest accuracy improvements but significantly reduces throughput, limiting its utility for latency-sensitive deployments. In contrast, 4-bit quantization provides the strongest memory savings but suffers from severe accuracy degradation, making it only suitable for highly constrained environments. FP16 remains the most balanced choice, preserving both inference speed and power efficiency, and thus is preferable in most practical scenarios.

From a deployment perspective, Locaris addresses several practical challenges. Real-world systems often face inconsistent access point availability and variable telemetry support, yet Locaris remains robust under these conditions. Moreover, it avoids the rigid preprocessing pipelines that baselines depend on, such as top-K AP selection, RSS standardization, and dataset-specific thresholds. Instead, it directly processes raw telemetry values and metadata, making it more flexible and generalizable across diverse deployment environments. This operational flexibility, combined with its ability to achieve sub-meter localization in ideal settings and sub-5 meter accuracy under degraded input, makes it particularly well-suited for large-scale indoor localization systems.

Training Locaris on larger telemetry datasets can poten-
tially make it even more accurate and improve its ability to generalize across diverse settings. We envision that Locaris can scale in two ways: either by building a full regression model from scratch or by adding compact adapter modules
to enable richer tasks like 3D localization or environment
mapping. Our experiments across different datasets showed
strong results both when trained separately and together,
proving that the model can learn patterns that transfer across
places. As long as it has enough relevant data, the method
stays flexible and reliable.

\section{Conclusion}

In this paper, we introduce Locaris, a multi-telemetry modality, LLM-based Wi-Fi indoor positioning system.

Rather than using embeddings or few-shot prompts, we treat indoor positioning as a next-token prediction task, aligning the output distribution with predicted distance as the final token. This approach leverages LLMs' autoregressive modeling, internal attention mechanisms, positional encodings, and emergent sequence forecasting behaviors, capturing sequential dependencies that static methods miss. Our simplified pipeline consists of a single autoregressive pass that is generalizable across sensors and environments.

Our experimental study of Locaris' performance using different datasets representing various environments shows it can reliably capture complex signal behavior like multipath without filtering or preprocessing, while consistently staying under 1m error. Our results demonstrate that token-based regression is a viable alternative to traditional methods - It is schema-less by design, requires no preprocessing pipeline, and effectively captures semantic relationships. It also highlights how LLMs can implicitly learn nuanced spatial patterns from noisy wireless telemetry by attending to structured input sequences and understanding the language of wireless.

\bibliographystyle{IEEEtran}
\bibliography{references}

\begin{thebibliography}{10}
\providecommand{\url}[1]{#1}
\csname url@samestyle\endcsname
\providecommand{\newblock}{\relax}
\providecommand{\bibinfo}[2]{#2}
\providecommand{\BIBentrySTDinterwordspacing}{\spaceskip=0pt\relax}
\providecommand{\BIBentryALTinterwordstretchfactor}{4}
\providecommand{\BIBentryALTinterwordspacing}{\spaceskip=\fontdimen2\font plus
\BIBentryALTinterwordstretchfactor\fontdimen3\font minus \fontdimen4\font\relax}
\providecommand{\BIBforeignlanguage}[2]{{%
\expandafter\ifx\csname l@#1\endcsname\relax
\typeout{** WARNING: IEEEtran.bst: No hyphenation pattern has been}%
\typeout{** loaded for the language `#1'. Using the pattern for}%
\typeout{** the default language instead.}%
\else
\language=\csname l@#1\endcsname
\fi
#2}}
\providecommand{\BIBdecl}{\relax}
\BIBdecl

\bibitem{9531633}
N.~Singh, S.~Choe, and R.~Punmiya, ``Machine learning based indoor localization using wi-fi rssi fingerprints: An overview,'' \emph{IEEE Access}, vol.~9, pp. 127\,150--127\,174, 2021.

\bibitem{farahsari2022survey}
P.~S. Farahsari, A.~Farahzadi, J.~Rezazadeh, and A.~Bagheri, ``A survey on indoor positioning systems for iot-based applications,'' \emph{IEEE Internet of Things Journal}, vol.~9, no.~10, pp. 7680--7699, 2022.

\bibitem{correa2017review}
A.~Correa, M.~Barcelo, A.~Morell, and J.~L. Vicario, ``A review of pedestrian indoor positioning systems for mass market applications,'' \emph{Sensors}, vol.~17, no.~8, p. 1927, 2017.

\bibitem{dwiyasa2016survey}
F.~Dwiyasa and M.-H. Lim, ``A survey of problems and approaches in wireless-based indoor positioning,'' in \emph{2016 International conference on indoor positioning and indoor navigation (IPIN)}.\hskip 1em plus 0.5em minus 0.4em\relax IEEE, 2016, pp. 1--7.

\bibitem{deak2012survey}
G.~Deak, K.~Curran, and J.~Condell, ``A survey of active and passive indoor localisation systems,'' \emph{Computer Communications}, vol.~35, no.~16, pp. 1939--1954, 2012.

\bibitem{twala2009empirical}
B.~Twala, ``An empirical comparison of techniques for handling incomplete data using decision trees,'' \emph{Applied Artificial Intelligence}, vol.~23, no.~5, pp. 373--405, 2009.

\bibitem{nessa2020survey}
A.~Nessa, B.~Adhikari, F.~Hussain, and X.~N. Fernando, ``A survey of machine learning for indoor positioning,'' \emph{IEEE access}, vol.~8, pp. 214\,945--214\,965, 2020.

\bibitem{roy2022survey}
P.~Roy and C.~Chowdhury, ``A survey on ubiquitous wifi-based indoor localization system for smartphone users from implementation perspectives,'' \emph{CCF Transactions on Pervasive Computing and Interaction}, vol.~4, no.~3, pp. 298--318, 2022.

\bibitem{10609490}
E.~Pagliari, L.~Davoli, and G.~Ferrari, ``Wi-fi-based real-time uav localization: A comparative analysis between rssi-based and ftm-based approaches,'' \emph{IEEE Transactions on Aerospace and Electronic Systems}, vol.~60, no.~6, pp. 8757--8778, 2024.

\bibitem{8692423}
F.~Zafari, A.~Gkelias, and K.~K. Leung, ``A survey of indoor localization systems and technologies,'' \emph{IEEE Communications Surveys \& Tutorials}, vol.~21, no.~3, pp. 2568--2599, 2019.

\bibitem{cominelli2023exposing}
M.~Cominelli, F.~Gringoli, and F.~Restuccia, ``Exposing the csi: A systematic investigation of csi-based wi-fi sensing capabilities and limitations,'' in \emph{2023 IEEE International Conference on Pervasive Computing and Communications (PerCom)}.\hskip 1em plus 0.5em minus 0.4em\relax IEEE, 2023, pp. 81--90.

\bibitem{beyer1999nearest}
K.~Beyer, J.~Goldstein, R.~Ramakrishnan, and U.~Shaft, ``When is “nearest neighbor” meaningful?'' in \emph{International conference on database theory}.\hskip 1em plus 0.5em minus 0.4em\relax Springer, 1999, pp. 217--235.

\bibitem{hegselmann2023tabllm}
S.~Hegselmann, A.~Buendia, H.~Lang, M.~Agrawal, X.~Jiang, and D.~Sontag, ``Tabllm: Few-shot classification of tabular data with large language models,'' in \emph{International conference on artificial intelligence and statistics}.\hskip 1em plus 0.5em minus 0.4em\relax PMLR, 2023, pp. 5549--5581.

\bibitem{pires2019multilingual}
T.~Pires, E.~Schlinger, and D.~Garrette, ``How multilingual is multilingual bert?'' \emph{arXiv preprint arXiv:1906.01502}, 2019.

\bibitem{google2024dolphingemma}
\BIBentryALTinterwordspacing
{Google DeepMind}, ``Dolphingemma: How google ai is helping decode dolphin communication,'' \url{https://blog.google/technology/ai/dolphingemma/}, February 2024, accessed: 2025-05-03. [Online]. Available: \url{https://blog.google/technology/ai/dolphingemma/}
\BIBentrySTDinterwordspacing

\bibitem{liu2025can}
H.~Liu, H.~Kamarthi, Z.~Zhao, S.~Xu, S.~Wang, Q.~Wen, T.~Hartvigsen, F.~Wang, and B.~A. Prakash, ``How can time series analysis benefit from multiple modalities? a survey and outlook,'' \emph{arXiv preprint arXiv:2503.11835}, 2025.

\bibitem{10.1145/3677846.3677854}
\BIBentryALTinterwordspacing
W.~Aljedaani, A.~Habib, A.~Aljohani, M.~Eler, and Y.~Feng, ``Does chatgpt generate accessible code? investigating accessibility challenges in llm-generated source code,'' in \emph{Proceedings of the 21st International Web for All Conference}, ser. W4A '24.\hskip 1em plus 0.5em minus 0.4em\relax New York, NY, USA: Association for Computing Machinery, 2024, p. 165–176. [Online]. Available: \url{https://doi.org/10.1145/3677846.3677854}
\BIBentrySTDinterwordspacing

\bibitem{hu2021loralowrankadaptationlarge}
\BIBentryALTinterwordspacing
E.~J. Hu, Y.~Shen, P.~Wallis, Z.~Allen-Zhu, Y.~Li, S.~Wang, L.~Wang, and W.~Chen, ``Lora: Low-rank adaptation of large language models,'' 2021. [Online]. Available: \url{https://arxiv.org/abs/2106.09685}
\BIBentrySTDinterwordspacing

\bibitem{MSIndoorLoc2021}
Y.~Hu, X.~Fan, Z.~Yin, F.~Qian, Z.~Ji, Y.~Shu, Y.~Han, Q.~Xu, J.~Liu, and P.~Bahl, ``{The Wisdom of 1,170 Teams: Lessons and Experiences from a Large Indoor Localization Competition},'' in \emph{Proceedings of the 29th Annual International Conference on Mobile Computing and Networking (MobiCom '23)}, 2023, pp. 1--15.

\bibitem{lymberopoulos2017microsoft}
D.~Lymberopoulos and J.~Liu, ``The microsoft indoor localization competition: Experiences and lessons learned,'' \emph{IEEE Signal Processing Magazine}, vol.~34, no.~5, pp. 125--140, 2017.

\bibitem{pagano2015indoor}
S.~Pagano, S.~Peirani, and M.~Valle, ``Indoor ranging and localisation algorithm based on received signal strength indicator using statistic parameters for wireless sensor networks,'' \emph{IET Wireless Sensor Systems}, vol.~5, no.~5, pp. 243--249, 2015.

\bibitem{mistry2015rssi}
H.~P. Mistry and N.~H. Mistry, ``Rssi based localization scheme in wireless sensor networks: A survey,'' in \emph{2015 Fifth International Conference on Advanced Computing \& Communication Technologies}.\hskip 1em plus 0.5em minus 0.4em\relax IEEE, 2015, pp. 647--652.

\bibitem{yiu2017wireless}
S.~Yiu, M.~Dashti, H.~Claussen, and F.~Perez-Cruz, ``Wireless rssi fingerprinting localization,'' \emph{Signal Processing}, vol. 131, pp. 235--244, 2017.

\bibitem{singh2021machine}
N.~Singh, S.~Choe, and R.~Punmiya, ``Machine learning based indoor localization using wi-fi rssi fingerprints: An overview,'' \emph{IEEE access}, vol.~9, pp. 127\,150--127\,174, 2021.

\bibitem{9733026}
V.~Barral~Vales, O.~C. Fernández, T.~Domínguez-Bolaño, C.~J. Escudero, and J.~A. García-Naya, ``Fine time measurement for the internet of things: A practical approach using esp32,'' \emph{IEEE Internet of Things Journal}, vol.~9, no.~19, pp. 18\,305--18\,318, 2022.

\bibitem{ibrahim2018verification}
M.~Ibrahim, H.~Liu, M.~Jawahar, V.~Nguyen, M.~Gruteser, R.~Howard, B.~Yu, and F.~Bai, ``Verification: Accuracy evaluation of wifi fine time measurements on an open platform,'' in \emph{Proceedings of the 24th annual international conference on mobile computing and networking}, 2018, pp. 417--427.

\bibitem{bullmann2020comparison}
M.~Bullmann, T.~Fetzer, F.~Ebner, M.~Ebner, F.~Deinzer, and M.~Grzegorzek, ``Comparison of 2.4 ghz wifi ftm-and rssi-based indoor positioning methods in realistic scenarios,'' \emph{Sensors}, vol.~20, no.~16, p. 4515, 2020.

\bibitem{qorib2024decoder}
M.~Qorib, G.~Moon, and H.~T. Ng, ``Are decoder-only language models better than encoder-only language models in understanding word meaning?'' in \emph{Findings of the Association for Computational Linguistics ACL 2024}, 2024, pp. 16\,339--16\,347.

\bibitem{koroteev2021bert}
M.~V. Koroteev, ``Bert: a review of applications in natural language processing and understanding,'' \emph{arXiv preprint arXiv:2103.11943}, 2021.

\bibitem{zhao2024mininglimiteddatasufficiently}
\BIBentryALTinterwordspacing
Z.~Zhao, F.~Meng, H.~Li, X.~Li, and G.~Zhu, ``Mining limited data sufficiently: A bert-inspired approach for csi time series application in wireless communication and sensing,'' 2024. [Online]. Available: \url{https://arxiv.org/abs/2412.06861}
\BIBentrySTDinterwordspacing

\bibitem{cesar2023bert}
L.~B. Cesar, M.-{\'A}. Manso-Callejo, and C.-I. Cira, ``Bert (bidirectional encoder representations from transformers) for missing data imputation in solar irradiance time series,'' \emph{Engineering Proceedings}, vol.~39, no.~1, p.~26, 2023.

\bibitem{ai2023gpt}
O.~AI, ``Gpt-4 technical report,'' \emph{arXiv preprint arXiv:2303.08774}, 2023.

\bibitem{touvron2023llama}
H.~Touvron, T.~Lavril, G.~Izacard, X.~Martinet, M.-A. Lachaux, T.~Lacroix, B.~Rozi{\`e}re, N.~Goyal, E.~Hambro, F.~Azhar \emph{et~al.}, ``Llama: Open and efficient foundation language models,'' \emph{arXiv preprint arXiv:2302.13971}, 2023.

\bibitem{hu2022lora}
E.~J. Hu, Y.~Shen, P.~Wallis, Z.~Allen-Zhu, Y.~Li, S.~Wang, L.~Wang, W.~Chen \emph{et~al.}, ``Lora: Low-rank adaptation of large language models.'' \emph{ICLR}, vol.~1, no.~2, p.~3, 2022.

\bibitem{hoffmann2022trainingcomputeoptimallargelanguage}
\BIBentryALTinterwordspacing
J.~Hoffmann, S.~Borgeaud, A.~Mensch, E.~Buchatskaya, T.~Cai, E.~Rutherford, D.~de~Las~Casas, L.~A. Hendricks, J.~Welbl, A.~Clark, T.~Hennigan, E.~Noland, K.~Millican, G.~van~den Driessche, B.~Damoc, A.~Guy, S.~Osindero, K.~Simonyan, E.~Elsen, J.~W. Rae, O.~Vinyals, and L.~Sifre, ``Training compute-optimal large language models,'' 2022. [Online]. Available: \url{https://arxiv.org/abs/2203.15556}
\BIBentrySTDinterwordspacing

\bibitem{zenedo}
\BIBentryALTinterwordspacing
X.~Feng, K.~A. Nguyen, and Z.~Luo, ``{WiFi RSS \& RTT dataset with different LOS conditions for indoor positioning},'' Zenodo, Jun. 2024, accessed: May 11, 2025. [Online]. Available: \url{https://zenodo.org/doi/10.5281/zenodo.11558791}
\BIBentrySTDinterwordspacing

\bibitem{bi2022supplementary}
\BIBentryALTinterwordspacing
J.~Bi, Y.~Wang, B.~Yu \emph{et~al.}, ``Supplementary open dataset for wifi indoor localization based on received signal strength,'' \emph{Satellite Navigation}, vol.~3, no.~1, p.~25, 2022. [Online]. Available: \url{https://github.com/bijingxue/SODIndoorLoc}
\BIBentrySTDinterwordspacing

\bibitem{feng2024wifi}
X.~Feng, K.~A. Nguyen, and Z.~Luo, ``A wifi rss-rtt indoor positioning system using dynamic model switching algorithm,'' \emph{IEEE Journal of Indoor and Seamless Positioning and Navigation}, 2024.

\bibitem{Optuna}
T.~Akiba, S.~Sano, T.~Yanase, T.~Ohta, and M.~Koyama, ``{Optuna: A Next-generation Hyperparameter Optimization Framework},'' \emph{arXiv preprint arXiv:1907.10902}, 2019.

\end{thebibliography}

\vspace{12pt}

\end{document}